\documentclass[letterpaper, 10 pt, conference]{ieeeconf}  % 

\IEEEoverridecommandlockouts                              % 
\usepackage{graphics} % for pdf, bitmapped graphics files
\usepackage{mathptmx} % assumes new font selection scheme installed
\usepackage{times} % assumes new font selection scheme installed
\usepackage{amsmath} % assumes amsmath package installed
\usepackage{amssymb}  % assumes amsmath package installed
\usepackage{xcolor}
\usepackage{graphicx} 
\usepackage{float} % Adicione se ainda não tiver, para usar o [H]
\usepackage{pifont}
\usepackage[hidelinks]{hyperref}
\usepackage{tabularx}
\usepackage{makecell}
\usepackage{ragged2e} % for \RaggedRight
\usepackage{booktabs} 
\usepackage{bbding}
\usepackage{threeparttable} % in the preamble

\title{\LARGE \bf
Why Personalization Matters: Cross-Subject Challenges in EMG-IMU-based HRI Activity Recognition}

\author{Ruan R. C. de F. Carminati$^{1,3}$, Giovanni Braglia$^{2}$, Luigi Biagiotti$^{3}$, \\ Ronnier F. Rohrich$^{1}$, André S. de Oliveira$^{1}$, Mikael N. Hartmann$^{1}$, and André E. Lazzaretti$^{1}$% <-this % stops a space
\thanks{The project is supported by the National Council for Scientific and Technological Development (CNPq) under grant number 407984/2022-4; the Fund for Scientific and Technological Development (FNDCT); the Ministry of Science, Technology and Innovations (MCTI) of Brazil; the Funding Authority for Studies and Projects (FINEP) under project number 0368/18; the Araucaria Foundation; the General Superintendence of Science, Technology and Higher Education (SETI); and NAPI Robotics.}
\thanks{$^{1}$Universidade Tecnológica Federal do Paraná, Curitiba, Brazil,
        {\tt\small ruan@alunos.utfpr.edu.br}}%
\thanks{$^{2}$Istituto Italiano di Tecnologia, Genoa, Italy.}
\thanks{$^{3}$University of Modena and Reggio Emilia, Emilia-Romagna, Italy.} }

\begin{document}

\maketitle
\thispagestyle{empty}
\pagestyle{empty}

\begin{abstract}

This paper investigates wearable-based recognition of human activities and gestures to support Human–Robot Interaction (HRI) in object-handover and assembly-like scenarios. Electromyography (EMG) and Inertial Measurement Unit (IMU) signals were collected using a Myo armband, culminating in a novel dataset introduced as MAGIC-HRI (Multimodal Activity, Gesture and Intention Collection) with a large taxonomy of 53 movement classes, including Brazilian Sign Language (LIBRAS) numbers (0–9), hand gestures, object/tool handover actions (pick up/give/hold), tool-manipulation tasks, and generic assembly/idle motions, collected from 11 participants with 10 samples per class (530 samples per participant). Signals are segmented by detecting muscle activation via an EMG energy envelope, then processed using sliding windows; time- and frequency-domain features are extracted. Multiple classical classifiers are tuned via cross-validated grid search, with Random Forest as the strongest baseline. A Leave-One-Subject-Out (LOSO) protocol reveals a large generalization gap, indicating substantial subject dependence. A personalized adaptation experiment suggests that injecting a small number of samples from a new user can markedly improve recognition. Overall, the study contributes a broad, HRI-driven multimodal dataset, a rigorous evaluation emphasizing generalization, and practical evidence that personalization is likely required for robust deployment in practical HRI.

\end{abstract}

%%%%%%%%%%%%%%%%%%%%%%%%%%%%%%%%%%%%%%%%%%%%%%%%%%%%
\section{Introduction}\label{sec:introduction}
%%%%%%%%%%%%%%%%%%%%%%%%%%%%%%%%%%%%%%%%%%%%%%%%%%%%

% Human–robot interaction (HRI) encompasses several important scenarios, including safe human–robot coexistence, adaptive motion planning, collaborative task execution, and direct physical interaction. Safety is a key concern, as robots must recognize both human presence and their own operational limits during shared tasks \cite{Scholz2025}. Among collaborative actions, object handover is particularly important because it requires precise synchronization between the human and the robot during grasping and release to prevent the object from being dropped \cite{Duan2024}. According to \cite{Ortenzi2021}, handover is a joint action between a giver and a receiver, comprising a pre-handover phase, in which intent is communicated through visual, spoken, or tactile cues \cite{Costanzo2021}, and a physical handover phase, corresponding to the actual transfer of the object. This makes human intention especially relevant, since recognizing intent helps identify critical moments in the interaction \cite{Khanna2026}.

Human-Robot Interaction (HRI) encompasses key scenarios like safe human–robot coexistence, adaptive motion planning, collaborative task execution, and direct physical interaction. Safety is crucial, as robots must recognize human presence and their own limits during shared tasks \cite{Scholz2025}. Among collaborative actions, object handover is particularly critical, requiring precise human-robot synchronization during grasping and release to prevent the object from being dropped. Handover is a joint giver-receiver action \cite{Ortenzi2021} comprising a pre-handover phase where intent is communicated via visual, spoken, or tactile cues \cite{Costanzo2021} and a physical phase for the actual object transfer. Consequently, human intention is highly relevant, as recognizing it identifies critical interaction moments \cite{Khanna2026}.

\begin{figure}[t]
    \vspace*{1.5mm} %
    \centering
    \includegraphics[width=1\linewidth]{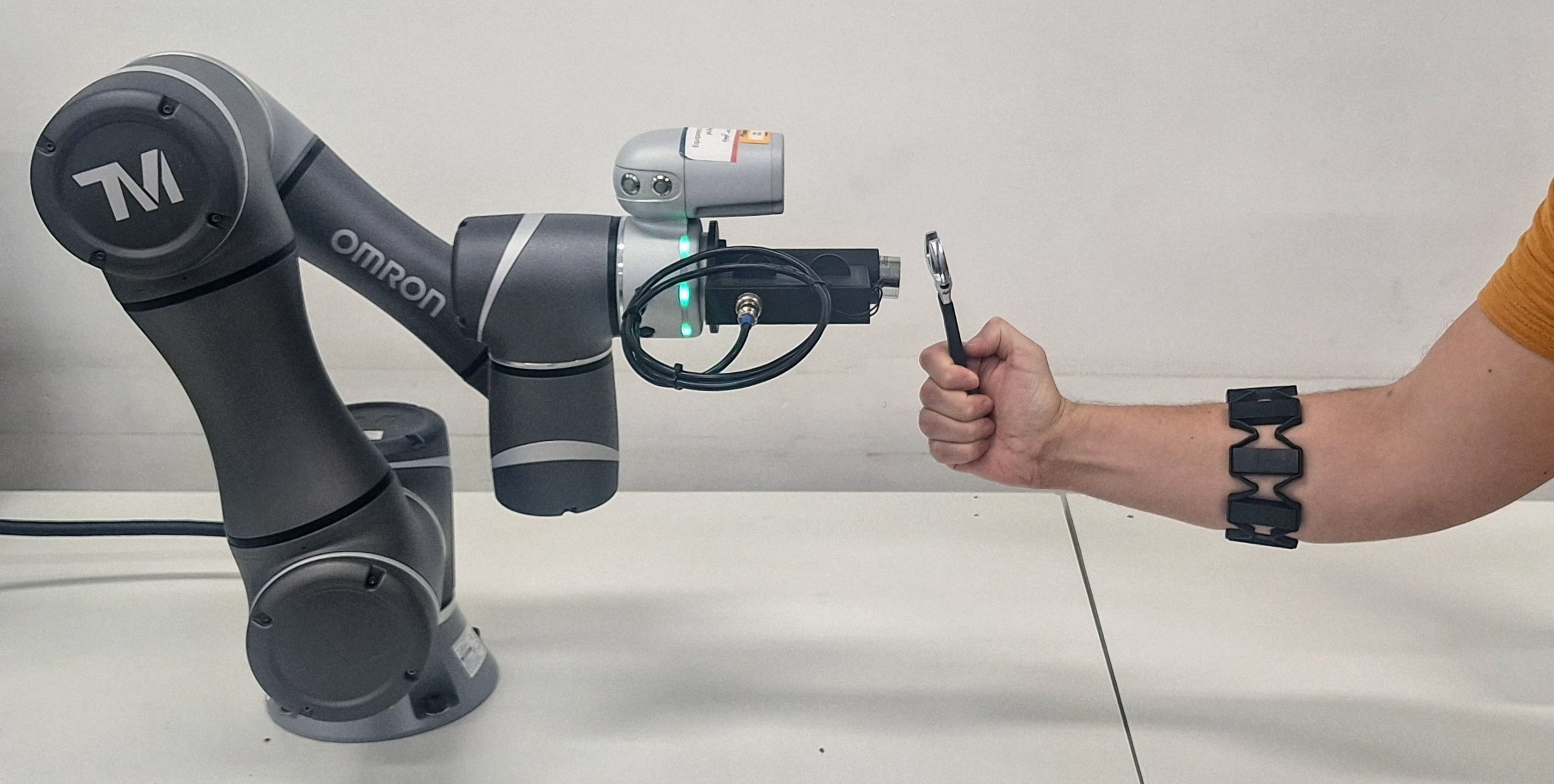}
    \caption{General EMG/IMU-based industrial HRI in our context.}
    \label{fig:ExperimentalSetup}
\end{figure}

Data-driven approaches are essential for interpreting human intention in HRI and enabling safer, more natural collaboration \cite{Rakholia2024}. Multiple sensing modalities can support this goal, each with specific strengths and limitations. Vision sensors provide contactless monitoring and often do not require markers, but they are affected by occlusion, poor lighting, privacy issues, and highly dynamic environments \cite{Zou2024, Dong2025, Wang2022}. Sound sensors enable intuitive, hands-free communication, though industrial noise can significantly degrade performance \cite{Wang2022, Dong2025}. Force/torque sensors improve safety during physical contact and are less dependent on environmental conditions, but they may struggle to distinguish intentional from unintentional contact and can be influenced by vibration and noise \cite{Zou2024}. In general, Electromyography (EMG) and Inertial Measurement Unit (IMU) sensors are a better choice for industrial HRI, the focus of this work (illustrated in Figure \ref{fig:ExperimentalSetup}), because they provide direct, privacy-preserving measurements of worker motion and muscular intention while remaining robust to occlusion, poor lighting, clutter, and other harsh factory conditions \cite{Khanna2026, Dong2025, Wang2019}.

EMG/IMU typically requires careful calibration and poses challenges for data interpretation for different actions. Particularly, the following questions can be identified as relevant in this context: (i) How reliably can wearable EMG and IMU sensor data be used to classify a large number of complex handover interactions in an industrial context? (ii) How does inter-subject variability affect the generalization capability of classification models when the system interacts with previously unseen human operators, i.e., Leave-One-Subject-Out (LOSO) analysis \cite{Gholamiangonabadi2020}? And (iii) What strategies are most effective in mitigating performance degradation and ensuring seamless interaction when the robotic system encounters a new user? 

Considering the aspects discussed so far, Table \ref{tab:comparison_related_works} shows that most prior HRI studies evaluate only a small number of highly distinct classes, such as 10 in \cite{Wang2022}, 6 in \cite{Wang2019}, 8 in \cite{Wang2025}, 3 in \cite{Du2024}, or only binary states in \cite{Sun2022}, which limits their realism for real settings. In contrast, the present study considers 53 distinct movement classes and relies on a wearable sensor, rather than cameras, eye-tracking, or vision-based pose extraction \cite{Sun2022, Cai2024, Zhao2022, Mitra2025}, making it more flexible and less sensitive to environmental conditions. While previous works investigating robotic interactions or datasets are typically constrained to just a few distinct actions, often relying on virtual environments \cite{Pettersson2023} or offline benchmarks \cite{Abidi2024}, this work bridges the gap between experimental realism and high-dimensional taxonomy by capturing actual handover dynamics across 53 movement classes during real interactions with a collaborative robot. Table \ref{tab:comparison_related_works} shows that the number of subjects in other studies is similar to this work. In addition, unlike studies based on standard cross-validation \cite{Dong2025, Wang2019}, which may overestimate performance by mixing data from the same subjects in training and testing, this work adopts a LOSO protocol, better exposing the true generalization challenge and motivating a personalization strategy based on sample inclusion.

\begin{table}[t]
\vspace*{1.5mm} %
\centering
\caption{Comparison of our proposed method with related works in terms of input data and classification scope.}
\label{tab:comparison_related_works}
\resizebox{\columnwidth}{!}{%
\begin{threeparttable}
\begin{tabular}{l c c c c c}
\hline
\textbf{Ref.} & \textbf{Input data} & \textbf{LOSO} & \textbf{Dataset} & \textbf{Classes}  &\textbf{Subjects}\\
\hline
\cite{Dong2025}       & Vision, EMG/IMU                & \XSolid        & \XSolid        & 18  &1\\
\cite{Wang2022}       & Vision, Sound, EMG/IMU         & \XSolid        & \XSolid        & 10  &10\\
 \cite{Wang2019}       & EMG/IMU                         & \XSolid        & \XSolid        & 9   &6\\
\cite{Wang2025}       & Vision, Force/Torque, EMG/IMU  & \XSolid        & \XSolid        & 8   &6\\
 \cite{Du2024}         & Vision, Force/Torque, EMG/IMU  & \XSolid        & \XSolid        & 3   &7\\
 \cite{Sun2022}        & Vision                         & \XSolid        & \XSolid        & 7   &-\\
\cite{Cai2024}        & Vision                         & \XSolid        & \XSolid        & 6   &9\\
 \cite{Zhao2022}       & Vision                         & \XSolid        & \XSolid        & 12  &-\\
 \cite{Mitra2025}      & Vision                         & \XSolid        & \XSolid        & 6   &13\\
 \cite{Pettersson2023} & Vision                         & \XSolid        & \XSolid        & 10  &21\\
\cite{Abidi2024}      & EMG/IMU                        & \XSolid        & \XSolid        & 12  &-\\
\hline
\textbf{Ours}         & EMG/IMU                        & \CheckmarkBold & \CheckmarkBold & \textbf{53}  &\textbf{11}\\
\hline
\end{tabular}
\begin{tablenotes}
\footnotesize
\item Note: \textit{LOSO} indicates whether LOSO evaluation was performed. \textit{Dataset} indicates whether the study publicly released its dataset. The \textit{Subjects} indicates the number of individuals whose data were collected in each study and subsequently used to train the model.
\end{tablenotes}
\end{threeparttable}%
}
\end{table}

Hence, the primary contribution of this paper is the introduction of a novel EMG/IMU dataset comprising 53 distinct classes, collected in an industrial application from 11 subjects. Also, the following additional contributions can be highlighted in this work:

\begin{itemize}
    \item An experimental benchmark evaluating the effectiveness of the 53-class dataset in predicting complex human intents for robotic control.
    \item A comprehensive cross-validation analysis to assess model generalization and practical applicability across subjects in real-world HRI scenarios.
    \item A methodological approach for evaluating scenarios in which a model is adapted for new users, outlining practical implications and strategies for maintaining high performance when integrating a new operator into the framework.
\end{itemize}

The rest of the paper is structured as follows: Section II describes the dataset construction, signal processing methods, and the evaluation protocol. Section III presents the results and evaluation. Section IV discusses the findings, and Section V provides concluding remarks.

%%%%%%%%%%%%%%%%%%%%%%%%%%%%%%%%%%%%%%%%%%%%%%%%%%%%
\section{Wearable-Based Activity and Gesture Recognition}\label{sec:wearable}
\begin{figure*}[htb]
    \vspace*{1.5mm} %
    \centering
    \includegraphics[width=0.9\textwidth]{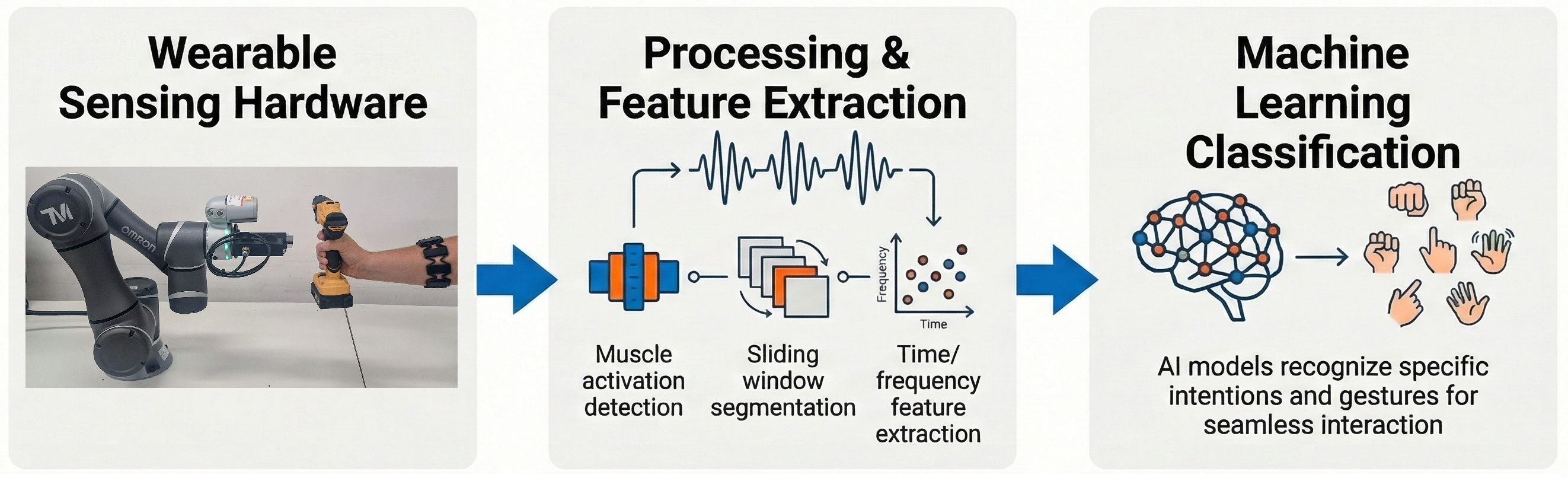}
    \caption{Overview of the wearable-based activity and gesture recognition pipeline evaluated in this work.}
    \label{fig:overview}
\end{figure*}

%%%%%%%%%%%%%%%%%%%%%%%%%%%%%%%%%%%%%%%%%%%%%%%%%%%%
The proposed methodology applies data collection, signal processing, and pattern recognition, as presented in Figure \ref{fig:overview}. The following sections will go in-depth on those topics, showing the processes for each step. The dataset\footnote{To collect the dataset with volunteers, we obtained approval from the Ethical Committee for Research in Humans of the Federal University of Technology-Paraná (CAAE 91430125.0.0000.0177). } is available at \url{https://github.com/ruancarminati/MAGIC-HRI-V01.git} (MAGIC-HRI: Multimodal Activity, Gesture, and Intention Collection for HRI).

%---------------------------------------------------
\subsection{Sensing Setup}\label{subsec:}
%---------------------------------------------------
The experimental setup consists of a wearable gesture-recognition interface and a collaborative robotic arm. The Myo Armband (Model MYOD5) from Thalmic Labs is a non-invasive device that collects data via eight EMG electrodes. It features IMU sensors, which include a 3D accelerometer and a 3D gyroscope. The system captures EMG samples at up to 200 Hz and IMU samples at up to 50 Hz. This device had been shown to be efficient in similar applications, as seen in \cite{Wang2019} and \cite{Mendes2020}. The OMRON TM5-700 from Techman Robot is a 6-axis collaborative robotic system. During data acquisition, the robot delivered objects to the subjects and also retrieved them.

%---------------------------------------------------
\subsection{MAGIC-HRI Dataset Description}\label{subsec:dataset}
%---------------------------------------------------

The MAGIC-HRI dataset was developed at the Applied Robotics Laboratory at UTFPR. Data collection included 11 participants (9 men and 2 women) with a mean age of 33 years (range: 22 to 54). The dataset consists of EMG and IMU signals recorded at 50 Hz as participants performed a series of predefined tasks and gestures.  At the beginning of the collection, the armband was placed on the right forearm, with the channel 3 electrode on the \textit{flexor carpi ulnaris} muscle \cite{Mendes2020}. All the participants were right-handed. A total of 53 movement classes were defined. Those movements were chosen to group a range of possible interactions between the robot and the user, using only the EMG/IMU sensor during the assembly process. Each participant contributed 10 samples per class, resulting in 530 samples and approximately 57 minutes of raw data per individual. The duration of each movement class varies depending on the class (range: 3 s to 30 s). The movement classes are categorized as follows:

\begin{itemize}
    \item \textbf{Numbers:} Cardinal numbers 0 through 9 represented in Brazilian Sign Language (LIBRAS). As the EMG/IMU armband is the unique source of communication, the numbers can then be used to execute more complex commands that simpler gestures wouldn't be able to represent. For example, help the robot pick one specific tool from many available options. 
    \item \textbf{Gestures:} Basic hand configurations such as spread fingers, open palm, wave in and out, thumb-to-middle, closed hand, and thumbs up. Those gestures are used to represent the most general/frequent commands.
    \item \textbf{Object Interaction:} Data collected for actions including holding, giving objects or tools to a robot, and picking up objects or tools from a robot.
    \item \textbf{Tool Manipulation:} Screwing and unscrewing tasks performed using an electric screwdriver, an adjustable wrench, or a Phillips screwdriver. And active use of tools, including a hot glue gun and a heat gun.
    \item \textbf{Generic Movements:} This category includes idle states, empty-hand movements, and general assembly tasks. Those movements were recorded, with the possibility of avoiding noise in the classification process in mind.
\end{itemize}

%---------------------------------------------------
\subsection{Signal Processing and Feature Extraction}\label{subsec:signal}
%---------------------------------------------------
% Detail all the steps until the feature extraction, including the features. 

Signal processing begins with the detection of muscle activation. The processes consist of calculating the average energy across all 8 EMG channels. The energy calculation process was inspired by \cite{Su2016}. First, the raw EMG signals from all eight channels are squared to compute the instantaneous power. Then, a 50 ms moving-average window is applied to the squared values to create a smooth energy envelope, which is then averaged across all channels to produce a single representative signal of total muscle activity. Subsequently, the segments in which the energy exceeds 10\% of the peak are identified. Finally, activations shorter than 150 ms are considered noise and are excluded.

Because standard classifiers struggle with variable-length inputs, a sliding-window with overlap approach \cite{Hakonen2015} was applied to address the varying sample sizes over time in the MAGIC-HRI dataset. Among the various windowing techniques available, this study utilizes an approach defined by a specific window duration and overlap percentage. The period represents the window size, and the overlap represents how much data is shared with other segments. For example, a sample of 2 s, with a period of 1 s and an overlap of 50\%. After segmentation, there will be 3 segments: 0 to 1 s, 0.5 to 1.5 s, and 1 to 2 s.

After segmentation, feature extraction is performed. In this stage, the raw data is transformed into useful information. The features are divided into two domains: time and frequency. In the time-domain, 10 features were extracted from EMG signals, and 8 from IMU signals. In the frequency-domain, before extracting features, it was necessary to apply Welch's Method to estimate the signal's Power Spectral Density (PSD). This is necessary to transform the data from the time domain to the frequency-domain. In this method, the signal is divided into smaller segments, and a Fast Fourier Transform (FFT) is performed on each segment to obtain the power spectrum. With this data, the frequency features were calculated. 5 features were extracted from the EMG signals, and 5 from the IMU signals. The feature extraction process and the selected features draw inspiration from \cite{Mendes2020}. Accordingly, the features extracted in this work are presented as follows, categorized by sensor and signal domain:

\begin{itemize}
    \item \textbf{Time Domain (EMG):} Root Mean Square (RMS), Mean Absolute Value (MAV), Zero Crossings (ZC), Waveform Length (WL), Slope Sign Changes (SSC), Willison Amplitude (WAMP), Difference Absolute Standard Deviation Value (DASDV), Variance (VAR), Log Detector (LOG), and Sample Entropy (SMPEN).
    \item \textbf{Frequency Domain (EMG):} Mean Frequency (MNF), Median Frequency (MDF), Peak Frequency (PKF), Spectral Entropy (SE), and Total Power (TP).
    \item \textbf{Time Domain (IMU):} Arithmetic Mean (MEAN), Standard Deviation (STD), Root Mean Square (RMS), Skewness (SKEW), Kurtosis (KURT), Zero Crossing Rate (ZCR), Waveform Length (WL), and Signal Energy (ENG).
    \item \textbf{Frequency Domain (IMU):} Mean Frequency (MNF), Median Frequency (MDF), Peak Frequency (PKF), Spectral Entropy (SE), and Total Power (TP).
\end{itemize}

%---------------------------------------------------
\subsection{Classification Methods}\label{subsec:classification}
%---------------------------------------------------

Based on the works from \cite{Wang2025}, \cite{Du2024}, and \cite{Mendes2020}, several established machine learning and deep learning classifiers were employed in this study. Parameter grids were constructed for each classifier to facilitate cross-validated grid search. Each search space included various combinations of functional parameters, such as the number of estimators, kernel types, and hidden layer architectures. A 5-fold cross-validation was performed to ensure that the resulting accuracy scores were statistically robust and not influenced by any single data split. Classifier performance varied across scenarios; therefore, multiple classifiers were evaluated to determine the most suitable models for this application. All models were evaluated based on their classification accuracy and F1-score across the 53 classes. The dataset was partitioned into 75\% for training and 25\% for testing, and all models were trained using a 1 s window with 50\% overlap as the baseline experimental setting. 

%---------------------------------------------------
\subsection{Training and Evaluation Protocol}\label{subsec:training}
%---------------------------------------------------

The performance of the classifiers was evaluated using accuracy and F1-score. Multiple scenarios were developed by varying model parameters. In addition to a 75/25 split of the MAGIC-HRI for training and testing, a LOSO evaluation was implemented. The dataset comprised data from 11 subjects; for each iteration, the model was trained on data from 10 subjects and tested on the remaining subject. This process was repeated for all subjects, and the average accuracy and F1-score were calculated.

%%%%%%%%%%%%%%%%%%%%%%%%%%%%%%%%%%%%%%%%%%%%%%%%%%%%
\section{Experimental Evaluation}\label{sec:experimental}
%%%%%%%%%%%%%%%%%%%%%%%%%%%%%%%%%%%%%%%%%%%%%%%%%%%%

Initially, a comparison of different classifiers was evaluated, following the procedure detailed in Subsection~\ref{subsec:classification}. According to Table \ref{tab:classifier_performance}, the best performance was achieved with random forest, using 150 estimators, a maximum depth of 30, a minimum sample split of 2, and a maximum number of features based on the square root. 

\begin{table}[t]
\vspace*{1.5mm} %
\centering
\small
\setlength{\tabcolsep}{8pt}
\caption{Accuracy performance of various classifiers across MAGIC-HRI class categories.}
\label{tab:classifier_performance}
\begin{tabular}{c c c}
\hline
\textbf{Classifier} & \textbf{Accuracy}&\textbf{F1-Score}\\
\hline
Random Forest& 74.73\%&74.95\%\\
Support Vector Machine& 70.61\%&69.91\%\\
 1D Convolutional Neural Network& 70.67\%&62.39\%\\
 Multi-Layer Perceptron& 63.83\%&63.35\%\\
Linear Discriminant Analysis& 51.33\%&50.57\%\\
Decision Tree& 49.07\%&48.71\%\\
 Long Short-Term Memory & 42.70\%&25.76\%\\
Naive Bayes& 33.51\%&32.15\%\\
\hline
\end{tabular}
\end{table}

Subsequently, the model’s performance on unseen data was assessed involving all 11 participants. The mean F1-scores were measured for both a 75/25 split and a LOSO protocol with the corresponding values reported in Table \ref{tab:overlap_comparison_transposed}. A discrepancy of more than 60\% was observed between these two evaluation settings, indicating poor cross-subject generalization and suggesting possible overfitting. To examine whether this limitation could be mitigated by increasing the amount of training data, additional experiments were conducted using window overlaps of 75\% and 90\%, while preserving the 1 s window length. Although a higher overlap improved average performance in the 75/25 split scenario, no comparable improvement was observed in the LOSO evaluation. These results indicate that simply increasing the number of overlapping samples is insufficient to address the model’s difficulty in generalizing to previously unseen subjects.

\begin{table}[t]
\vspace*{1.5mm} %
\centering
\small
\setlength{\tabcolsep}{10pt}
\caption{Model performance metrics across different overlap configurations.}
\label{tab:overlap_comparison_transposed}
\begin{tabular}{c c c}
\hline
\textbf{Overlap} & \textbf{F1-Score}&\textbf{F1-Score LOSO}\\
\hline
50\% Overlap & 81.17\%&20.29\%\\
75\% Overlap & 91.15\%&20.19\%\\
90\% Overlap & 99.20\%&19.87\%\\
\hline
\end{tabular}
\end{table}

% To address overfitting, a feature-reduction strategy was investigated to improve model performance. The original model was trained using a total of 198 features. A SHapley Additive exPlanations (SHAP) analysis was performed to identify the most relevant features for the classifier \cite{Lundberg2017} and subsequently remove the features with the least importance. SHAP is a model-agnostic interpretability method that explains a machine learning model's output by assigning contribution scores to each feature, based on Shapley values from cooperative game theory. Before executing a SHAP analysis, it is necessary to remove highly correlated features, retaining only one representative feature per group. While there is no standard threshold, following the approach of \cite{Wei2025}, features with a correlation coefficient greater than 0.7 were excluded, reducing the initial 198 features to 74. The SHAP analysis was performed using a 1-second window with 90\% overlap on a Random Forest model with 50 estimators. Subsequently, the Random Forest was retrained using the parameters identified via grid search, incorporating feature filters based on the top-ranked SHAP values. These results are presented in Figure \ref{fig:PerfShap}. Despite the reduction in the number of features, the results indicated no significant improvement in model performance.

\begin{figure}[t]
    \centering
    \includegraphics[width=1\linewidth]{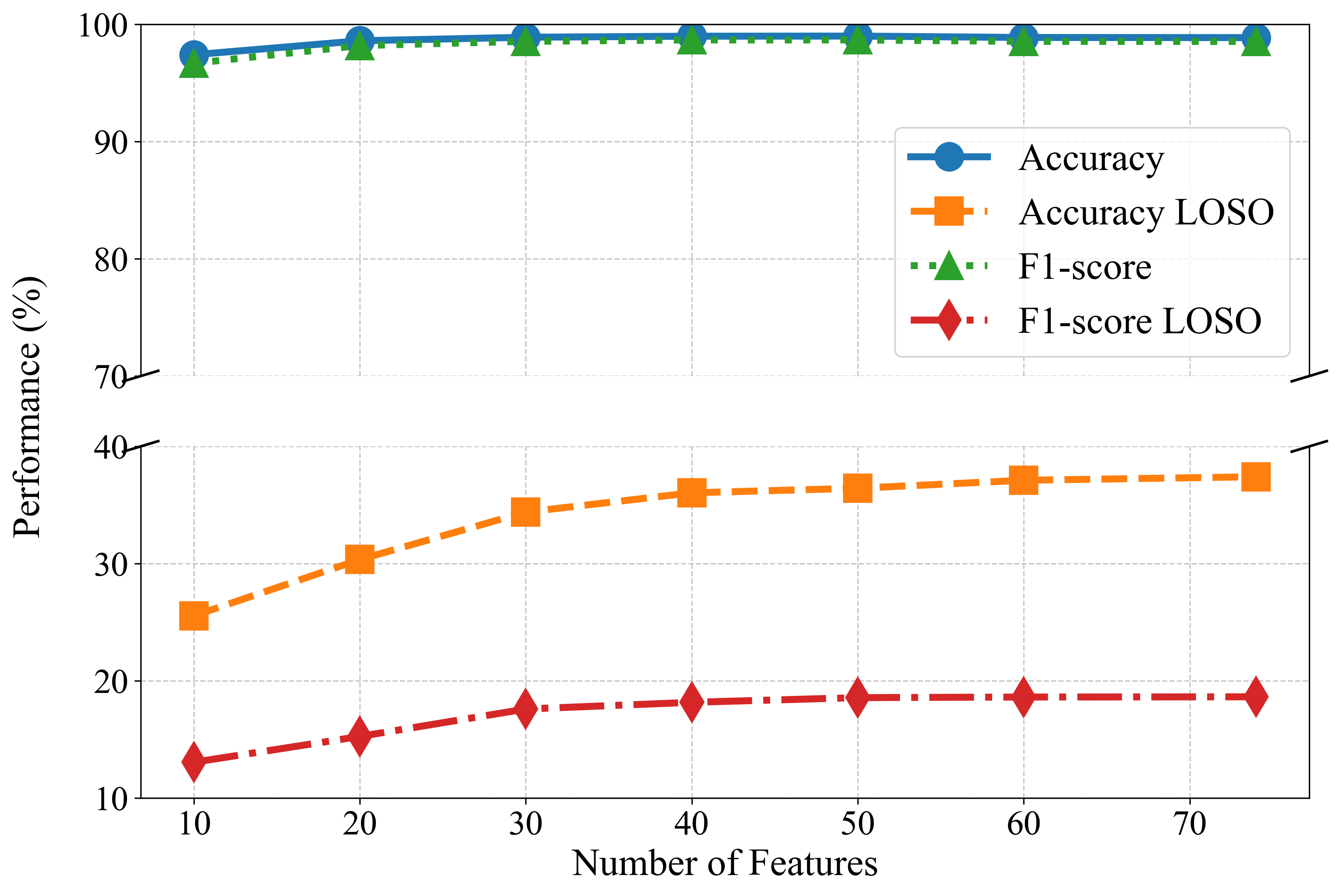}
    \caption{Model performance as a function of feature reduction based on SHAP values.}
    \label{fig:PerfShap}
\end{figure}

To address overfitting, a feature-reduction strategy was investigated. The original model was trained using 198 features. A SHapley Additive exPlanations (SHAP) analysis was performed to identify the most relevant features \cite{Lundberg2017} and remove those of least importance. SHAP is a model-agnostic interpretability method that explains a model's output by assigning contribution scores to each feature, based on Shapley values from cooperative game theory. Before executing a SHAP analysis, it is necessary to remove highly correlated features, retaining only one representative per group. Following the approach of \cite{Wei2025}, features with a correlation coefficient greater than 0.7 were excluded, reducing the initial 198 features to 74. The SHAP analysis was performed using a 1 s window with 90\% overlap on a Random Forest model with 50 estimators. Subsequently, the model was retrained using parameters identified via grid search, incorporating feature filters based on the top-ranked SHAP values. These results are presented in Figure \ref{fig:PerfShap}. Despite the reduction in features, the results indicated no significant improvement in model performance.

Building upon these findings and aligning with the best way to perform HRI, a personalized setup was considered to enable more effective interaction with unseen subjects. To assess this proposition, an experiment was conducted to estimate the number of gesture samples required for the system to adapt to a previously unseen subject. Since the dataset contains 10 samples per gesture per subject, 2 samples from each movement were reserved as an external test set to prevent data leakage. The LOSO protocol was extended by introducing an incremental adaptation loop that incorporates 1 additional sample from the left-out subject into the training set at each round. Accordingly, the procedure began with 0 samples from the excluded subject and progressed incrementally to 8, added cumulatively in chronological order. In all cases, only 40 features were used, as indicated by the SHAP analysis. To determine the effect of personalization on system performance, four testing scenarios are implemented:

% \textcolor{red}{Returning to the central objective of this work, namely handover between a robot and a human, a personalized setup was considered as a possible strategy to enable more effective interaction with new subjects. To assess this hypothesis, an experiment was conducted to estimate the number of gesture samples required for the system to adapt to and recognize signals from a previously unseen subject. Since the dataset contains 10 samples per gesture per subject, two samples from each movement were reserved as an external test set to prevent data leakage. The LOSO protocol was then extended by introducing an incremental adaptation loop, in which one additional sample from the left-out subject was incorporated into the training set at each round. Accordingly, the procedure began with no samples from the excluded subject and progressed incrementally to a total of eight samples. Samples are added cumulatively to the training set in chronological order of collection. In all cases, the number of features was fixed from the SHAP analysis to 40, as in our previous result. In order to determine the effect of personalization on system performance, four different testing scenarios are implemented: }

\begin{enumerate}
    \item \textbf{Internal Test (General):} The general internal test is performed using a 75/25 split of the initial training data pool.
    
    \item \textbf{Internal Test (Incremental Migration):} This scenario evaluates the model's adaptation by progressively adding samples from the unseen subject to the training set. In turn, the number of internal tests for a particular subject is reduced proportionally from eight to none samples, as more data from that subject is incorporated into the training process.
    
    \item \textbf{External Test (General):} A general evaluation based on the two samples reserved from all subjects except for the one being tested in this LOSO round. These samples have remained completely isolated from the training process and can be used as a measure of how well the system generalizes.
    
    \item \textbf{External Test (Target Subject):} A more specific evaluation based on only the two samples from the subject being adapted for this LOSO round. It is used as a measure of how well a person will be recognized on completely new data.
\end{enumerate}

\begin{figure}[t]
    \vspace*{1.5mm} %
    \centering
    \includegraphics[width=1\linewidth]{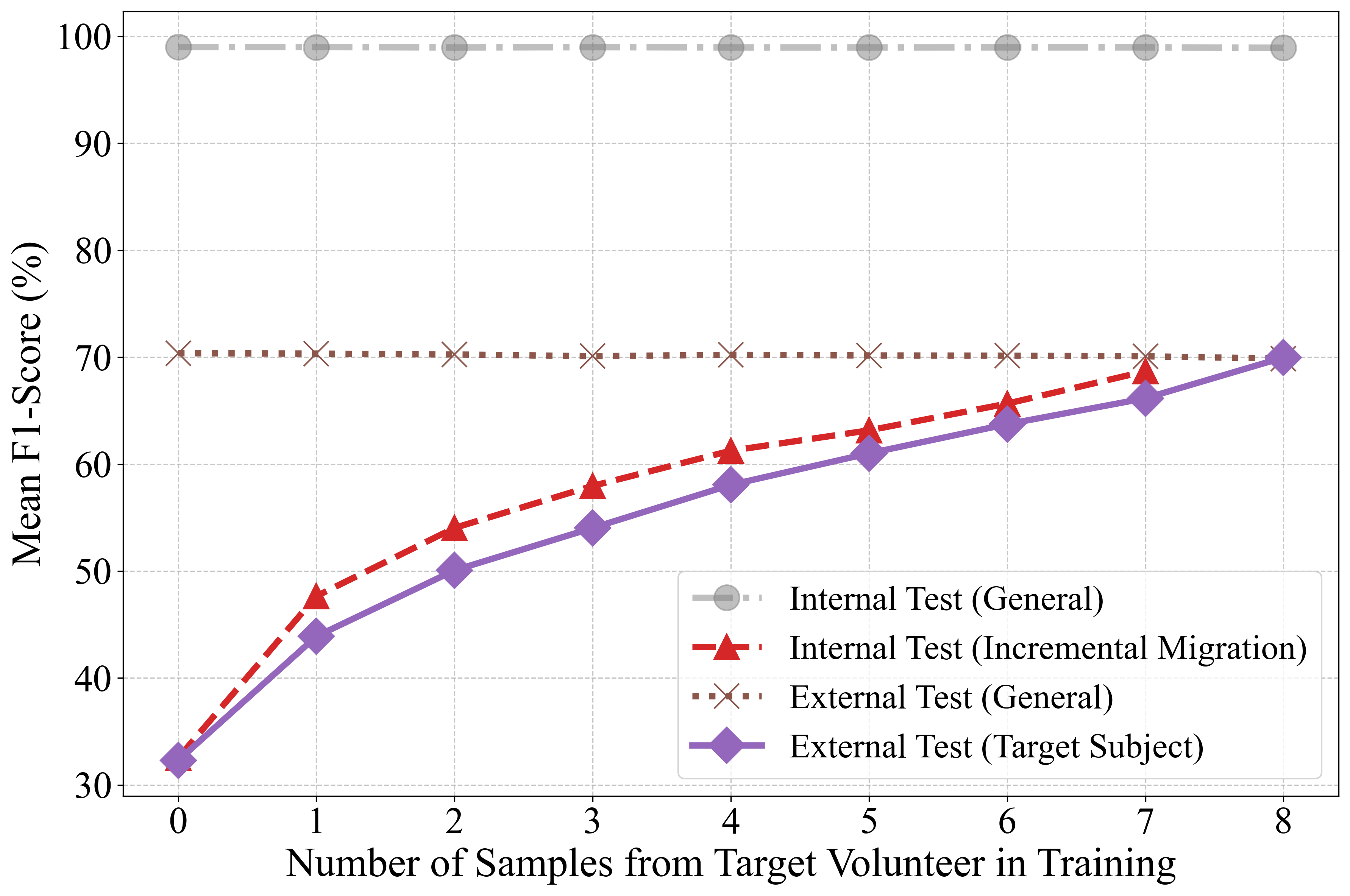}
    \caption{Impact of sample injection volume on LOSO F1-Score.}
    \label{fig:numberinjectionimg}
\end{figure}

As shown in Figure \ref{fig:numberinjectionimg}, there is a significant improvement in performance as more new samples are added to the model. This shows that personalization is a promising approach for successful human-robot handover using the current database. Although the performance is not yet at the baseline level established by the 75/25 split, the model recognizes more than 65\% of the signals once at least 7 personalization samples are added.

%%%%%%%%%%%%%%%%%%%%%%%%%%%%%%%%%%%%%%%%%%%%%%%%%%%%
\section{Discussion}\label{sec:discussion}
%%%%%%%%%%%%%%%%%%%%%%%%%%%%%%%%%%%%%%%%%%%%%%%%%%%%

In this work, we explored the challenges encountered in human–robot collaborative assembly tasks under high class counts and LOSO scenarios, which remain underexplored in the literature. In \cite{Wang2019}, the study most closely related to ours, the authors reported accuracies above 90\% across all six evaluated classes in both online and offline evaluations, using the same six subjects in both phases. Their classes focused on recognizing human handover intentions, enabling the human to control the handover process. Table~\ref{tab:comparisons} compares the accuracy of our approach (with and without LOSO) against \cite{Wang2019}. To align the datasets for this comparison, the most similar classes from MAGIC-HRI were selected, all classes of the ``give" type were merged into a single class, and random undersampling was applied to match the minority-class size.

\begin{table}[t]
\vspace*{1.5mm} %
\centering
\small
\setlength{\tabcolsep}{10pt}
    \caption{General comparisons with \cite{Wang2019}.}
    \begin{tabular}{cccc}
\hline
\textbf{Class}  & \textbf{Ref. \cite{Wang2019}} & \textbf{Ours} & \textbf{Ours (LOSO)}\\
\hline
Need& 93.33\%& 97.99\%& 62.90\%\\
       Give& 94.44\%& 90.22\%& 85.61\%\\
       Stop& 90.00\%& 93.55\%& 33.85\%\\
         Continue& 93.33\%& 94.11\%& 46.84\%\\
         Speed up& 93.33\%& 92.99\%& 67.59\%\\
         Slow down& 90.00\%& 97.72\%& 57.98\%\\
         \hline
    \end{tabular}
    \label{tab:comparisons}
\end{table}

Beyond considering a substantially smaller number of classes than in our work, the classes in \cite{Wang2019} correspond to actions that are clearly distinct from one another, as evidenced by accuracies above 90\% and comparable performance to our work when LOSO is not included. Moreover, the authors did not discuss in detail the performance obtained with subjects excluded from training, which makes a comparison in a LOSO setting difficult. This further reinforces the relevance of the present work and highlights important research directions in this field, especially for more complex HRI scenarios.

On the other hand, some limitations of the present approach remain, particularly given that this is ongoing research. One important limitation is the lack of a more in-depth robot-in-the-loop evaluation, that is, an assessment of classification performance during actual HRI in an online setting. Regarding computational feasibility, initial tests showed no hardware bottlenecks during inference, supporting its potential integration into real-time experiments. Overall, the LOSO results still present limitations for practical use, despite the evaluations and observations presented. In this sense, injecting personalized samples could increase the model's accuracy, as demonstrated here, thereby improving overall performance. Moreover, gathering more data across new subjects could help mitigate overfitting.

%%%%%%%%%%%%%%%%%%%%%%%%%%%%%%%%%%%%%%%%%%%%%%%%%%%%
\section{Conclusions and Future Work}\label{sec:conclusion}
%%%%%%%%%%%%%%%%%%%%%%%%%%%%%%%%%%%%%%%%%%%%%%%%%%%%

This paper advances wearable sensing for HRI by introducing and systematically evaluating MAGIC-HRI, a new 53-class EMG/IMU dataset designed for object handover and assembly-oriented industrial scenarios. Unlike prior HRI studies that typically address a small number of highly distinct classes, this work demonstrates a substantially broader recognition setting using data collected from 11 subjects during real interactions with a collaborative robot. Beyond the dataset itself, the paper presents a complete recognition pipeline based on muscle-activation segmentation, time- and frequency-domain feature extraction, and classical machine learning and deep learning classification, along with a rigorous benchmark across multiple classifiers. Among the evaluated models, Random Forest provided the strongest baseline performance, showing that feature-based wearable learning is viable for complex HRI intention and gesture recognition in realistic settings. 

% A second central contribution of this work is the explicit analysis of cross-subject generalization, which reveals that strong split-based results can substantially overestimate real deployment performance. The LOSO experiments show that subject variability is the main bottleneck for practical use and that neither increased overlap nor SHAP-based feature reduction may be sufficient to overcome it. Most importantly, the paper provides clear evidence that personalization is an effective deployment strategy: as subject-specific samples are progressively injected into training, recognition improves markedly, and the model can recognize more than 65\% of the signals once at least 7 personalization samples are added. This finding is highly relevant to real-world collaborative robotics, as it shows that a brief user calibration stage can significantly improve the reliability of wearable HRI recognition. To conclude, the main contributions of this study are the introduction of a large and realistic wearable HRI dataset, a rigorous evaluation framework centered on subject generalization, and a practical demonstration that personalization can bridge the gap between laboratory performance and robust real-world operation. Future work should expand the dataset, add more subjects, investigate models better suited to cross-subject transfer, and validate the approach in online robot-in-the-loop experiments.

A second central contribution is the explicit analysis of cross-subject generalization, revealing that strong split-based results can substantially overestimate real deployment performance. The LOSO experiments show that subject variability is the main bottleneck for practical use, and neither increased overlap nor SHAP-based feature reduction may be sufficient to overcome it. Most importantly, the paper provides clear evidence that personalization is an effective deployment strategy: as subject-specific samples are progressively injected into training, recognition improves markedly, allowing the model to recognize over 65\% of signals once at least 7 samples are added. This finding is highly relevant to real-world collaborative robotics, showing that a brief user calibration stage significantly improves the reliability of wearable HRI recognition.

To conclude, the main contributions are the introduction of a large, realistic wearable HRI dataset, a rigorous evaluation framework centered on subject generalization, and a practical demonstration that personalization bridges the gap between laboratory performance and robust real-world operation.

Future work should expand the dataset by adding more subjects and possibly new movements, investigate models better suited to cross-subject transfer, and validate the approach in online robot-in-the-loop experiments.

\section*{Acknowledgments}
The authors used Grammarly and Gemini 3.1 Pro to improve the grammar and overall quality of the English text.

\bibliographystyle{IEEEtran}
\bibliography{HandoverBibliography}

\end{document}